\documentclass{article}
\usepackage{ijcai21}

\usepackage{times}

\usepackage{soul}
\usepackage{url}
\usepackage[hidelinks]{hyperref}
\usepackage[utf8]{inputenc}
\usepackage[small]{caption}
\usepackage{graphicx}
\usepackage{amsmath}
\usepackage{amssymb}
\usepackage{booktabs}
\usepackage{fancyhdr}
\def\makeheadbox{{%
		\vspace{-3em}
		\hbox to0pt{\vbox{\baselineskip=10dd\hbox
				to\hsize{\kern3pt\vbox{\kern3pt
						\hbox{ \footnotesize \hspace{6em} This work was accepted at the 1\textsuperscript{st} International Workshop on Adaptive Cyber Defense (ACD) at the}
						\hbox{\footnotesize \hspace{8em} 30\textsuperscript{th} International Joint Conference on Artificial Intelligence (IJCAI-21)  August 19-20, 2021}
						\kern3pt}\hfil\kern3pt}}%
			\hss}}}

\makeatletter
\newcommand*\titleheader[1]{\gdef\@titleheader{#1}}
\AtBeginDocument{%
	\let\st@red@title\@title
	\def\@title{%
		\bgroup\normalfont\large\centering\@titleheader\par\egroup
		\vskip1.5em\st@red@title}
}
\makeatother

\title{A Smart and Defensive Human-Machine Approach to Code Analysis}

\titleheader{\vspace{-6em} \footnotesize  This work was accepted at the 1\textsuperscript{st} International Workshop on Adaptive Cyber Defense (ACD) at the 30\textsuperscript{th} International Joint Conference on Artificial Intelligence (IJCAI-21) August 19-20, 2021}

\author{
Fitzroy D. Nembhard\footnote{Contact Author}\and
Marco M. Carvalho
\affiliations
L3Harris Institute for Assured Information,\\ Florida Institute of Technology, Melbourne, Florida, USA\\
\emails
\{fnembhard, mcarvalho\}@fit.edu
}

\begin{document}

\maketitle
\begin{abstract}
Static analysis remains one of the most popular approaches for detecting and correcting poor or vulnerable program code. It involves the examination of code listings, test results, or other documentation to identify errors, violations of development standards, or other problems, with the ultimate goal of fixing these errors so that systems and software are as secure as possible. There exists a plethora of static analysis tools, which makes it challenging for businesses and programmers to select a tool to analyze their program code. It is imperative to find ways to improve code analysis so that it can be employed by cyber defenders to mitigate security risks. In this research, we propose a method that employs the use of virtual assistants to work with programmers to ensure that software are as safe as possible in order to protect safety-critical systems from data breaches and other attacks. The proposed method employs a recommender system that uses various metrics to help programmers select the most appropriate code analysis tool for their project and guides them through the analysis process. The system further tracks the user's behavior regarding the adoption of the recommended practices.
\end{abstract}

\section{Introduction}

The human element is a significant component of cybersecurity. Throughout the software development life cycle (SDLC) to the proper use of systems, humans are intrinsically involved in ensuring that systems are designed and used with security in mind. At the root of every network and computing system is the software that facilitates their operation. Consequently, to achieve certain guarantees of security, the underlying program code must be free of vulnerabilities. One way to eliminate vulnerabilities is through constant static analysis and bug fixing.  As researchers and engineers continue to create and improve code analysis tools, it is also essential to develop intelligent models that work with humans to inculcate behaviors that adaptively defend against attackers who may possess knowledge of system properties or security practices. Studies show that code-level vulnerabilities remain a top cause of data breaches across the globe \cite{synopsys_2021_ossra,cisq_software_quality_2020,software_fail_watch_2018}. In the 2021 Open Source Security Risk Analysis (OSSRA) report published by Synopsys, it was reported that 84\% of 1,546 codebases that were scanned contained at least 1 vulnerability \cite{synopsys_2021_ossra}. Nearly half of the open source vulnerabilities found in that report were identified as high risk \cite{synopsys_2021_ossra}. Given the fact that an attacker only knows what is perceived through observation of the target network, it is imperative to engineer methods to harden networks and the software that drive them to make it harder for unauthorized users to gain access. Consequently, defending a network from authorized access should start at the ground level\textemdash the program code.

Static analysis is a widely recommended approach to scan program code for unsafe practices. However, many programmers are skeptical of using static analysis tools due to the sheer number of tools available as well as usability challenges and poor presentation of vulnerability reports \cite{nembhard2019_analyzer_usability,nembhard_recommender_journal}. To address some of these limitations, researchers have proposed hybrid tools that combine static and dynamic approaches \cite{Aggarwal_integrating_static_dynamic,nembhard2018_dissertation} or use data fusion to aggregate and refine reports from a number of static analysis tools to provide the user with a more refined and comprehensive set of action items \cite{nembhard2018_dissertation,2016recoframework,isa_data_fusion_tool}. However, these methods have their own limitations in that they do not address the performance of the tools they employ.

While scanning code for vulnerabilities is an important step towards creating more secure software and networks, fixing the vulnerabilities found is a more critical activity. To that end, several researchers have proposed models that attempt to auto-fix or repair code \cite{sapfix_facebook_2019,gupta2017deepfix,survey_code_repair_2013}. These auto-repair systems have not yet become widespread due to their  ability to fix only a small subset of errors. There is also a need for systems that work with programmers to harden their code instead of simply scanning and displaying bug reports. We  were the first to apply recommender systems to the secure coding problem by providing the user with a set of candidates from which they can choose to fix vulnerabilities in their code \cite{nembhard_recommender_journal}. We used hand-selected features to train a model to detect taint-style (data dependence) vulnerabilities in program code. Our system, named VulIntel (Vulnerability Intellisensor), monitors program code as the programmer types, checks for unsafe practices, and then makes recommendations accordingly. We ranked the fixes using similarity metrics in order to provide the user with code that is most similar to the one being developed.   In addition, we  proposed  a system that  employs the use of virtual assistants to guide programmers as they scan their code for vulnerabilities \cite{nembhard2021conversational}. The system features a plug-and-play (PnP) model that can use \textit{any} code analyzer to scan a given project. The system was tested with Google Assistant and  PMD code analyzer.  The results demonstrate a seamless coordination of man and machine in an effort to encourage programmers to produce secure code.

In this position paper, we propose further enhancements to the model discussed in \cite{nembhard2021conversational}. The proposed improvements include adding a recommender system that is enriched with knowledge regarding existing static analysis tools,  building behavioral models that provide insights into the kinds of fixes that programmers choose, and tracking how users respond to a virtual agent that guides them in making defensive decisions regarding the security of their code and networks. The knowledge base will include information regarding static analysis tools such as trends and reviews, tool descriptions, types of projects targeted, etc, in order to help the programmer select the best tool to scan their project.  The ultimate goals are: 1) to increase the adoption of static analysis tools in the SDLC by teaming up users with a virtual assistant and recommender system to make code analysis more intelligent and human-centered, 2) to capture data that will help us understand how humans respond to secure recommendations, and 3) to foil insider threats that may be introduced during coding. The rest of the paper is organized as follows: Section \ref{sec:background} presents background information on techniques that will be employed in the proposed model. We discuss our proposed approach in Section \ref{sec:approach} and present our conclusion in Section \ref{conclusion}.

\section{Background}
\label{sec:background}
In this section, we discuss briefly some existing approaches that can be used to enrich a recommender system with knowledge about static analyzers as well as making the system more adaptive in its selection of a tool for a given project. 


\subsection{Topic Models}
A topic model is a generative model for documents, which specifies a probabilistic procedure by which documents can be
generated \cite{steyvers_probabilistic_topic_models}. Inversely, topic models can be used as an unsupervised machine learning technique to automatically cluster word groups and similar expressions that best characterize a set of documents.
Existing research has shown that topic models can be mined from source code \cite{nembhard_hybrid_2017,linstead2007mining}. 
Latent Dirichlet Allocation (LDA) is one example of a topic model that has been known to perform well in this domain \cite{linstead2007mining}. This is due to the fact that programmers tend to use informative keywords to name variables, methods, and classes, in addition to comments that describe the purpose of their code \cite{nembhard_hybrid_2017,Mirakhorli_tactics}.
These latent topics can be used to provide information on the overall focus or goal of a project.

\subsection{Topic Memory Networks for Short Text Classification}
\cite{zeng2018topic_networks} proposed Topic Memory Networks (TMN) for short text classification (for example, tool descriptions, reviews, etc). The proposed network consists of three major components:  1) a neural topic model (NTM), which is  based on a variational auto-encoder \cite{kingma2013auto}, to induce latent topics, 2) a topic memory mechanism that maps the inferred latent topics to classification features, and 3) a text classifier, which produces the final classification labels for instances \cite{zeng2018topic_networks}.

Given $X = {x_1, x_2, \ldots, x_M}$ as the input with $M$ short text instances, each instance $x$ is processed into two representations: 1 ) a bag-of-words (BoW) term vector $x_{BoW} \in \mathbb{R} ^V$ and 2) a word index sequence vector $x_{Seq} \in \mathbb{R}^L$ , where $V$ is the vocabulary size and $L$ is the sequence length. $x_{BoW}$ is fed into the neural topic model to induce latent topics.
The NTM is described in a manner similar to LDA, where each instance $x$ has a topic mixture $\theta$ that is represented as a $K$-dimensional distribution. Each topic $k$ is represented by a word distribution $\phi k$ over the given vocabulary. However, unlike LDA, in TMN, an encoder estimates the prior parameters and a decoder describes the generation story while the additional distributional vectors $\theta$ and $\phi$ yield latent topic representations \cite{zeng2018topic_networks}. Thus, topic memory networks are useful for understanding the purpose of a tool given a short review or description.

\section{Proposed Human-Machine Approach to Code Analysis}
\label{sec:approach}
\begin{figure}[!h]\centering
	\includegraphics[width=0.9\columnwidth]{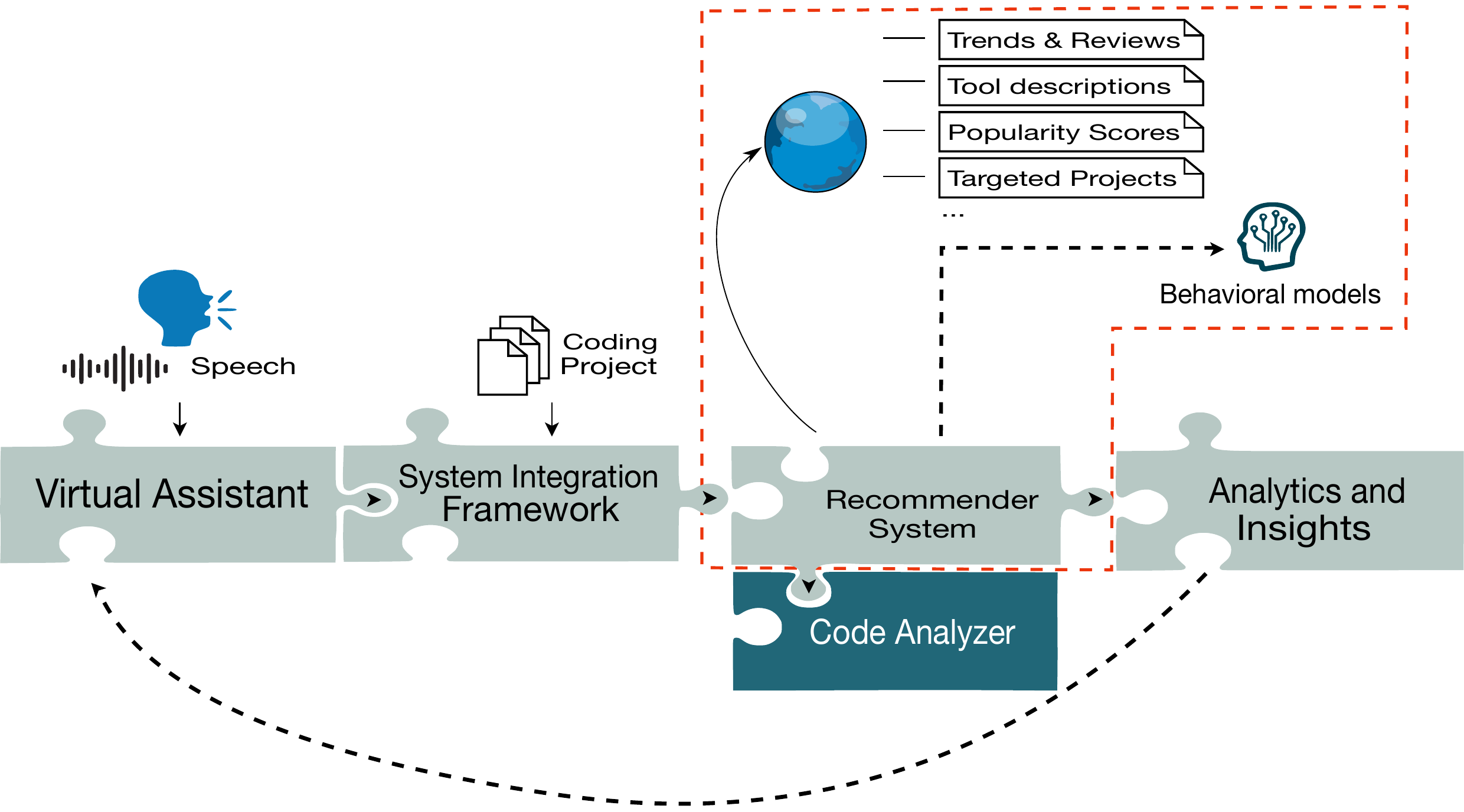}
	{\caption{Proposed human-machine approach to code analysis with enhancements highlighted using a dashed border.\label{fig:human_machine_approach}}}
\end{figure}

In this section, we describe the proposed approach for teaming humans with a virtual assistant to interactively scan and fix vulnerabilities in program code.
Figure \ref{fig:human_machine_approach} shows the architecture of the proposed framework.
The system has 5 main components: a virtual assistant, a system integration framework, a code analyzer, a recommender system, and an analytics and insights component.  The existing framework, as described in \cite{nembhard2021conversational}, involves creating and incorporating a Google Assistant app with a system integration framework that allows a user to conversationally scan a coding project for vulnerabilities. The user communicates with the virtual assistant using their voice, and web services backed by Google conversation APIs and local IDE plugins are used to contextualize the user's request to determine which project the user would like to scan. The system integration framework then uses a code analyzer to scan the system under test for unsafe code. Currently, the user may work with the virtual assistant to summarize vulnerabilities found and even email the user a vulnerability scan report.    

The additions proposed in this work are highlighted with a dashed border in Figure \ref{fig:human_machine_approach}. We now describe key components in the framework and the proposed improvements.

\subsection{The Machine: A Virtual Assistant}

A virtual assistant, or voice assistant, is a software agent that uses voice recognition, language processing algorithms, and voice synthesis to listen to specific voice commands and return relevant information or perform specific tasks on behalf of the user \cite{alan_blog_assistant_definition}. Voice commands are usually converted to text and processed by cloud-based services to determine the user's intent. Once intent has been determined, natural language processing (NLP) algorithms backed by machine learning and AI are used to  map the user's request to the most applicable function. Google Assistant was chosen as the virtual assistant for this research due to its popularity and easy-to-use App Engine \cite{appengine} and Dialogflow \cite{dialogflow} platforms. Google App Engine is a cloud computing platform as a service (PaaS) for developing and hosting web applications.   Dialogflow is a natural language understanding platform that allows users to design and integrate a conversational user interface into a mobile app, web application, device, bot, interactive voice response system, etc. \cite{dialogflow}. 

\begin{figure}[!h]\centering
	\includegraphics[width=0.9\columnwidth]{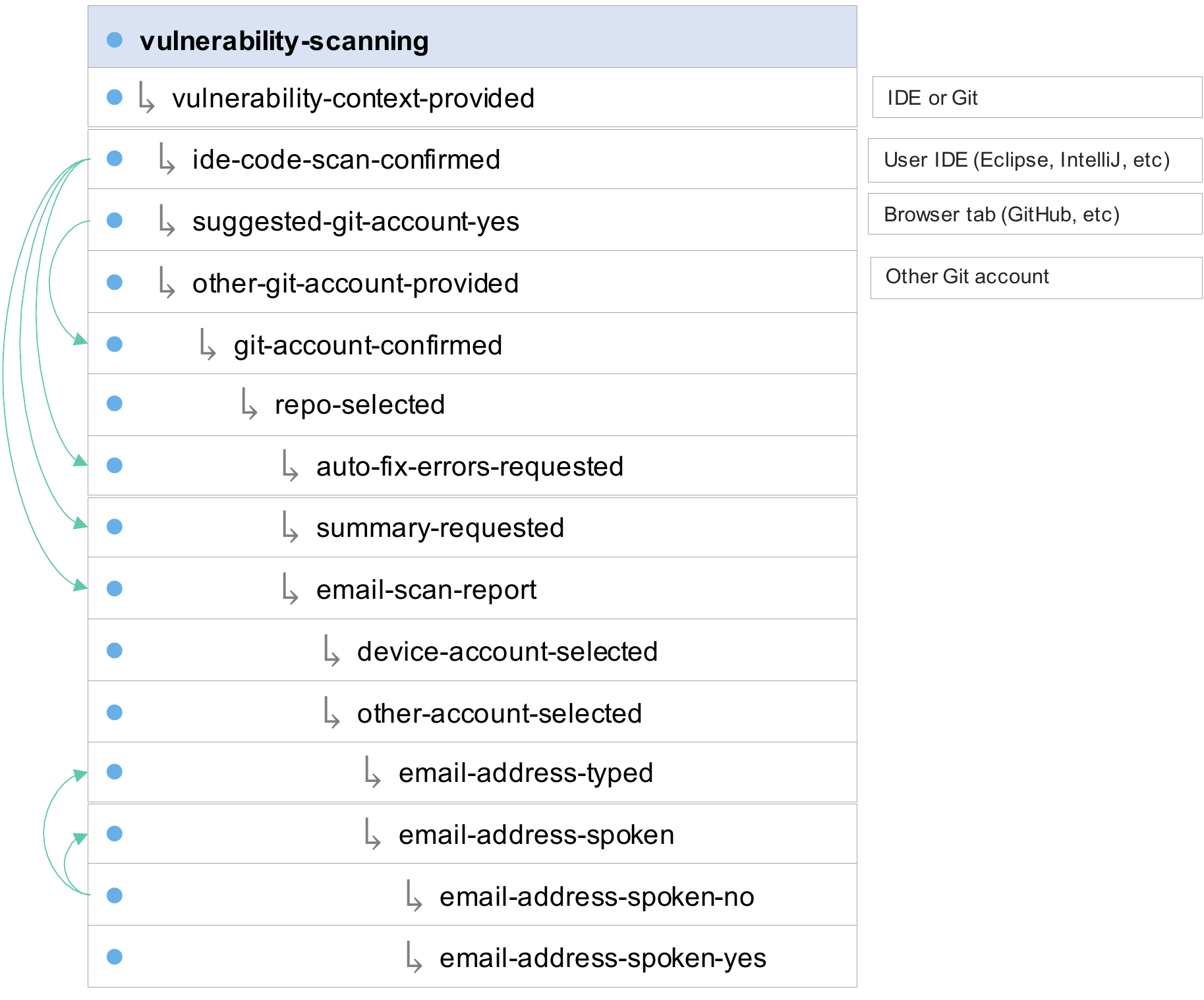}
	{\caption{Subset of Dialogflow intents already incorporated into the framework\label{fig:dialogflow_intents}}}
\end{figure}

The purpose of the virtual assistant is to work synchronously with the user to perform code analysis in a user-friendly environment. As a cloud-based agent, the virtual assistant can carry out tasks in a speedy and efficient manner as the user multitasks. Traditional code analysis frameworks may require users to manually configure system preferences or specify certain parameters in order for the analysis tool to work effectively. This may reduce the programmer's productivity and discourage use of static analysis tools altogether.  Our goal is to automate many of these manual tasks by allowing the agent to perform tasks on behalf of the user. Figure \ref{fig:dialogflow_intents} captures a subset of the intents already incorporated into the system \cite{nembhard2021conversational}. As shown in the diagram, the agent can work with the user to select a repository or coding project to scan, trigger the code analyzer to perform the scan, and summarize the results found.  More intents will be added to make the system more robust, active and responsive in the way it works with humans.

\subsection{Achieving Adaptability using a Recommender System}
Our existing framework \cite{nembhard2021conversational} can only use one code analyzer.  We intend to integrate a recommender system with the model so that the system can adaptively work with humans to select the most appropriate code analyzer. Information retrieval and text mining techniques will be used to collect data to enrich the knowledge base of the recommender system by including performance metrics and descriptions of existing tools so that it can recommend a code analyzer for a given project based on factors such as security, project size, project type, etc. Recommending a tool for a given project will make the system more adaptable to the unique security challenges facing a given institution, thus enabling experts to better ward off threats to their cyber-infrastructure. 

Adaptability can be achieved by fortifying the recommender system with both hand-selected and automatically selected data.  Most static analysis tools have websites that include short descriptions of the types of projects/errors that the tool targets. The Gartner Magic Quadrant \cite{gartner_magic_quadrant} and Gartner Peer Insights \cite{gartner_peer_insights} also include customer reviews and ratings for several static analysis tools that can provide a wealth of information for the recommender system.  Topic models and topic networks, which were discussed in Section \ref{sec:background}, will be used to map tools to topics and source code to topics in order to recommend the most appropriate tool for a given project. In other words,  topic networks will be used to augment topic models so the recommender system can suggest the most appropriate tool for a project even when very little information about a tool is provided.

\subsection{Behavioral Models}
Humans play an important role in the overall security and wellbeing of any computing system. The National Security Agency (NSA) lists \textquotedblleft understanding and accounting for human behavior\textquotedblright\ as one of the five hard problems of cybersecurity \cite{nsa_2019_5_hard_probs}. However, many designers of code analysis systems do not take into account the behaviors of humans regarding proper use and adoption of code analyzers. As users interact with the virtual assistant and their coding environment, we intend to collect data that can be used to model and understand human behavior in their response to code analysis. This kind of data will help us identify actions taken by users that lead to weaknesses in their coding projects. We also plan to use severity scores of vulnerabilities to determine potential impact on software security especially when recommendations are ignored by users. These models will also help to identify insider threats where programmers may intentionally reject secure practices and leave systems open to attacks. 

\begin{figure}[!h]\centering
	\includegraphics[width=0.9\columnwidth]{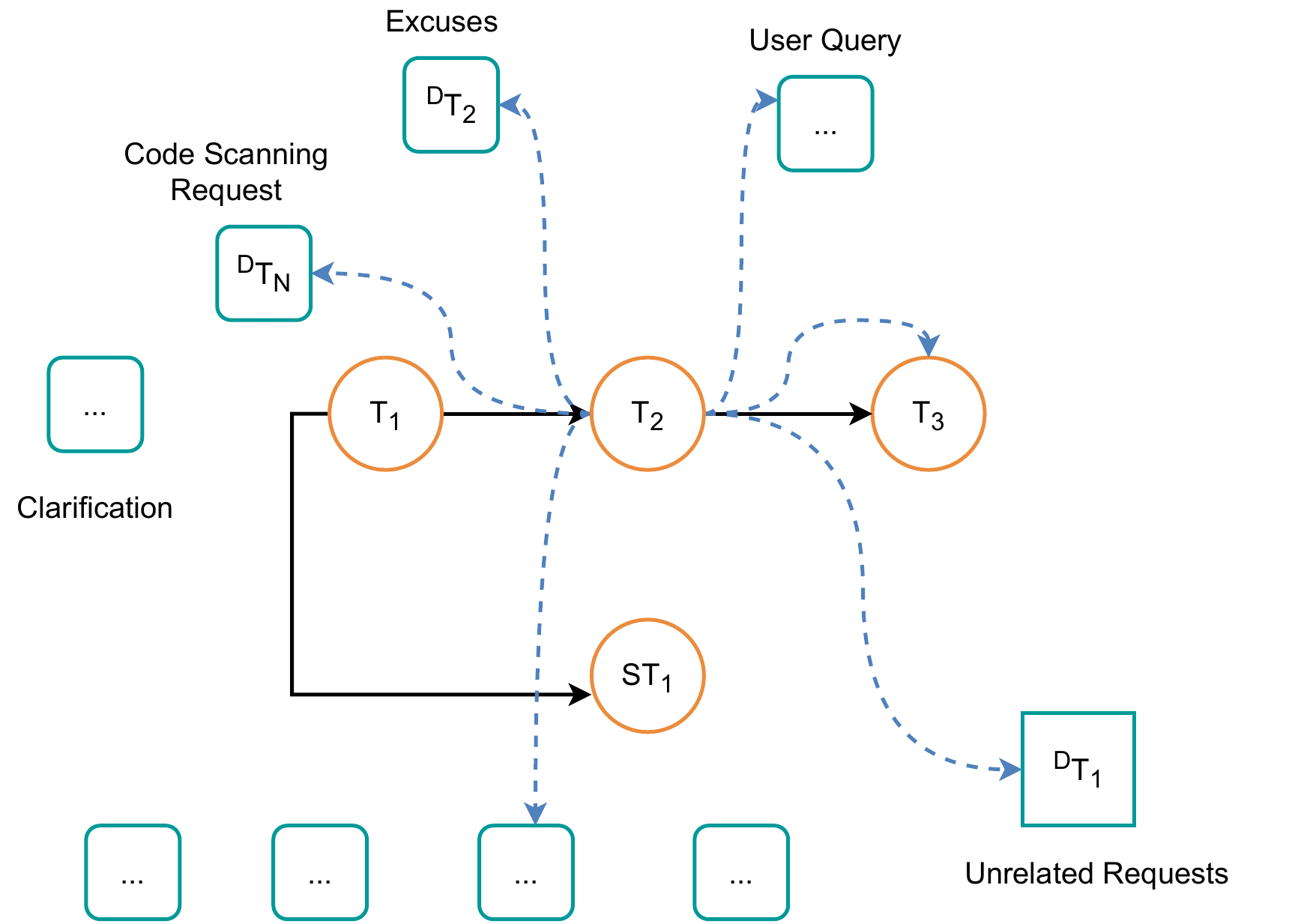}
	{\caption{An example state machine showing the dynamic nature of  conversations between a user and a virtual assistant\label{fig:code_analysis_fsm_states}}}
\end{figure}

Modeling context and diverse speech with Dialogflow is currently limited as the designer has to manually enumerate all user intents using a tree-like structure. This limitation can be observed in Figure \ref{fig:dialogflow_intents}. Further, once a user breaks the flow of a conversation, Dialogflow does not memorize the states or trail taken by the user. Research shows that attention and memory can be modeled using Long-short-term memory (LSTM) networks, which are a specific recurrent neural network (RNN) architecture that was designed to model temporal sequences and their long-range dependencies more accurately than conventional RNNs \cite{chen2019_graphflow,sak_google_2014_lstm,hochreiter1997_lstm}. In our research, we will explore the feasibility of integrating LSTMs with Dialogflow intents to handle more complex state-switching to make the conversation more realistic. Figure \ref{fig:code_analysis_fsm_states} shows an example state machine that models several paths that could be taken as users interact with the system. For example, a user may not like the results provided by a certain code analyzer and give an excuse or ask that a different tool be used. LSTMs will facilitate such types of deviations in the conversation.

\begin{figure}[!h]\centering
	\includegraphics[width=0.9\columnwidth]{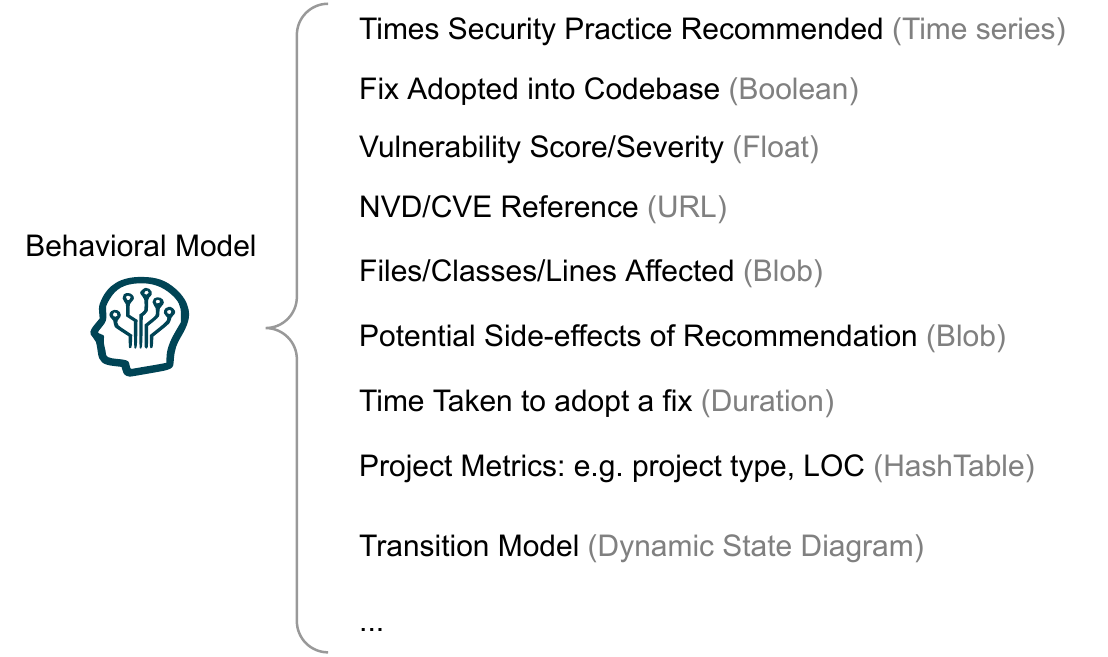}
	{\caption{Subset of proposed behavioral features to be captured by the recommender system \label{fig:code_analysis_behavioral_model}}}
\end{figure}

The path followed by the user as they communicate with the assistant will be used to create a behavioral model that will provide insights into how users use the code analysis framework. Figure \ref{fig:code_analysis_behavioral_model} shows preliminary features, including data-types, that can be used to describe human behavior as it relates to the adoption of safe practices during coding. Some example features include a time series of the times that a particular recommendation was given for a project, a Boolean value to indicate whether the recommendation was adopted into the codebase, and the transition model that represents the steps followed by the user in the analysis process, to name a few.  Behavioral models will be created using machine learning techniques such a ensemble methods and neural networks. We also intend to use feature selection to determine the most appropriate features based on the collected data.

\section{Conclusion and Future Work}
\label{conclusion}
In this position paper, we proposed enhancements to an existing human-centered code scanning framework that will make it more resilient in mitigating vulnerabilities and adaptive against insider threats. The existing system employs a virtual assistant that works with programmers to scan their code for vulnerabilities. The current system, while novel, is limited in that it uses a fixed static analysis tool and it does not adaptively defend against insider threats. 

Our subsequent plans are to implement the enhancements proposed in this paper. These enhancements include the addition of a recommender system that recommends and uses a code analyzer based on the user's preferences and the system under test. The model will be enriched with information such as tool descriptions, trends and reviews, target projects, etc, which will allow us to use advances in topic modeling to make appropriate recommendations. Further, as programmers use the proposed system, it will build behavioral models  that will help analysts spot insider threats and take necessary actions to eliminate or discourage practices that may expose the network and its underlying software to unauthorized  and malicious users. The system will be evaluated using user studies in the form of A/B testing where participants will use various tools to scan their code and report on their experiences.
\bibliographystyle{named}
\bibliography{ijcai21_smart_code_analysis}
\end{document}